\pdfoutput=1

\documentclass[11pt]{article}
\usepackage{pifont}
\usepackage[]{acl2023}
\usepackage{array}
\usepackage{multirow}
\usepackage{graphicx}
\usepackage{comment}
\usepackage{booktabs}
\usepackage{times}
\usepackage{latexsym}
\usepackage{enumitem}
\usepackage{subcaption}
\newcommand{\dataset}{{D3CODE~}}

\usepackage[T1]{fontenc}

\usepackage[utf8]{inputenc}

\usepackage{microtype}

\usepackage{inconsolata}

\title{D3CODE: Disentangling Disagreements in Data across Cultures\\ on Offensiveness Detection and Evaluation}

\author{Aida Davani \\
  Google Research\\
  \texttt{aidamd@google.com} \\\And
  Mark Díaz \\
  Google Research \\
  \texttt{markdiaz@google.com} \\ \And
  Dylan Baker \\
  DAIR Institute\\
  \texttt{dylan@dair-institute.org} \\ \AND
  Vinodkumar Prabhakaran \\
  Google Research \\
  \texttt{vinodkpg@google.com}
  }

\begin{document}
\maketitle
\begin{abstract}
\begin{itemize}
While human annotations play a crucial role in language technologies, annotator subjectivity has long been overlooked in data collection. Recent studies that have critically examined this issue are often situated in the Western context, and solely document differences across age, gender, or racial groups. As a result, NLP research on subjectivity have overlooked the fact that individuals within demographic groups may hold diverse values, which can influence their perceptions beyond their group norms.
To effectively 
incorporate these considerations into NLP pipelines, we need datasets with extensive parallel annotations from various social and cultural groups.
In this paper we introduce the \dataset dataset: a large-scale cross-cultural dataset of parallel annotations for offensive language in over 4.5K sentences annotated by a pool of over 4k annotators, balanced across gender and age, from across 21 countries, representing eight geo-cultural regions.
The dataset contains annotators' moral values captured along six moral foundations: care, equality, proportionality, authority, loyalty, and purity. 
Our analyses reveal substantial regional variations in annotators' perceptions that are shaped by individual moral values, offering crucial insights for building pluralistic, culturally sensitive NLP models.
\end{itemize}
\end{abstract}

\section{Introduction}
Designing Natural Language Processing (NLP) tools for detecting offensive or toxic text has long been an active area of research \cite{wulczyn2017ex,founta2018large}. However, applying traditional NLP solutions have led to overlooking the cultural and individual factors that shape humans' varying perspectives and disagreements on what is deemed offensive \cite{aroyo2015truth,talat2016you, salminen2019online,uma2021learning,prabhakaran2021releasing}. Perceiving language as offensive can depend inherently on one's moral judgments as well as the social norms dictated by the socio-cultural context within which one's assessments are made \cite{eickhoff2018cognitive,aroyo2019crowdsourcing,talat2021disembodied,rottger2021two,davani2023hate}. Therefore, data curating and modeling efforts should appropriately handle such subjective factors in order to better capture and learn human perspectives about offensiveness.


\begin{figure}
    \centering
    \includegraphics[width=.47\textwidth]{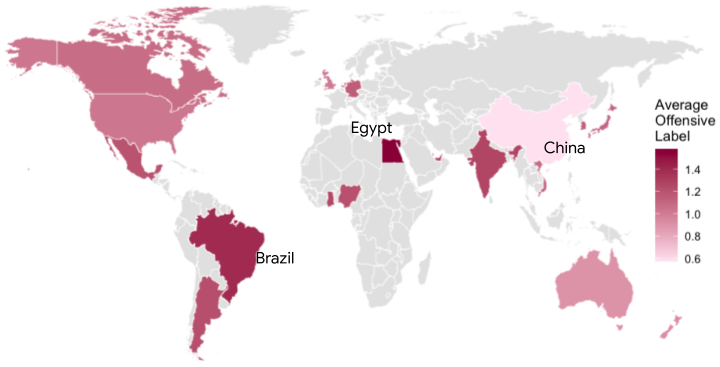}
    \caption{The distribution of labels provided from different countries. Annotators from China, Brazil, and Egypt provided significantly different labels.}
    \label{fig:countries}
\end{figure}

As a result, recent efforts call for diversifying the rater pools as well as designing models that look beyond predicting a singular ground truth \cite{davani2022dealing,aroyo2023reasonable}. However, the efforts for diversifying annotator pools often risk reducing annotators' differences to demographic variations. Moreover, subjectivity is often studied in relation to annotators' gender and race, 
in particular, within the Western context.
In reality, perceptions of what is offensive extend far beyond mere differences in demographics, shaped by an individual's lived experiences, cultural background and other psychological factors.

For instance,
the intricate interplay of social media content moderation and principles of freedom of speech brings the task of offensive language detection into the realm of moral and political deliberation (instances of such discussions can be found in \citet{balkin2017digital}, \citet{brannon2019free}, and \citet{kiritchenko2021confronting}). 
More generally, individuals might systematically disagree on notions of offensiveness, reflecting the complexity of beliefs and values that shape their perspectives and judgments within any given cultural context.
Therefore, we argue that the high divergence in annotators' perceptions of offensiveness \cite{prabhakaran2021releasing} can be traced back to individuals' diverse moral values along with 
the cultural and social norms that dictate the boundaries of acceptable language within a society.

In this work we introduce the \textbf{\dataset} dataset, built through a cross-cultural annotation effort aimed at collecting perspectives of offensiveness from 4309 participants of different age and genders across 21 countries within eight larger geo-cultural regions. Through an in-depth analysis of our dataset, we shed light on cultural and moral values that sets people apart during the annotation.
We believe that this dataset can be used for assessing modeling approaches that are designed to incorporate annotators' subjective views on language, as well as for evaluating different models' cultural and moral alignment.



\section{Related Work}

Recent studies have shown that treating annotators as interchangeable is not an effective approach for dealing with subjective language understanding tasks \citep{pavlick2019inherent, diaz2022crowdworksheets, prabhakaran2021releasing, davani2023hate}. Alternatively, modeling the nuances encoded in annotations and inter-annotator disagreements has recently been explored as an alternative solution for subjective tasks.

\subsection{Disagreement-aware Modeling}

When datasets include a set of annotations per instance, the distribution of these labels, and the disagreement extracted from the set, become two possible pieces of information that potentially help the modeling process.
\citet{basile2021need} argue that disagreement --- even on objective tasks \cite{parrish2023picture} --- should be considered as a source of information rather than being resolved. \citet{rottger2021two} propose a descriptive annotation paradigm for operationalizing subjectivity when surveying and modeling different beliefs. 

Therefore, incorporating inter-annotator agreements into the modeling process has gained more attention in the NLP community: \citet{plank2014learning} considered the item-level agreement as the loss function weights and achieved improvements on the downstream tasks. \citet{fornaciari2021beyond} leveraged annotator disagreement as an auxiliary task to be predicted along with ground-truth labels, which improves the performance even in less subjective tasks such as part-of-speech tagging. 

\citet{kennedy2020constructing} apply an item response theory model to the variations in annotations of hate speech to decompose the binary labels and use them in a multi-task model for predicting the latent variables.
While these methods intend to consider the variations in annotators' perspectives, they still fall short on regarding the integrity of the labels provided by each annotator and aggregate their varying subjectivities into a single construct.

\subsection{Annotator-aware Modeling}
The social nature of language means that social groups and relations play meaningful roles in how individuals use language, such as offensive speech \cite{diaz2022accounting}.
Acknowledging the differences in annotators' perceptions of subjective tasks has led model designers to consider information at the annotator level as the social factors needed for contextualizing language \citep{hovy2021importance}.  \citet{hovy2015demographic} show that providing the age or gender of the authors to text classifiers consistently and significantly improves the performance over demographic-agnostic models.
\citet{garten2019incorporating} model users' responses to questionnaire items based on their demographic information by training a demographics embedding layer, which can further be used in isolation to generate embeddings for any unseen sets of demographics.

\citet{ferracane-etal-2021-answer} add annotators' sentiment about the writer of the text into modeling their labels. They show that incorporating contextual information about annotators increases the performance. \citet{davani2022dealing} introduce a multi-annotator architecture that models each annotators' perspectives separately using a multi-task approach. 
And \citet{orlikowski-etal-2023-ecological} extend the multi-task model to capture perspectives of different groups, although they argued against modeling annotator groups. 
While these methods model annotations based on annotators' differences they do not incorporate psychological profile of annotators into modeling their perceptions of language, which are impacted by individual psychological traits, experiences, cultural background, and cognitive abilities.

\subsection{Annotators in NLP Datasets}
Although attending to annotators' background is gaining more attention, documenting how annotators' identity, which shapes their comprehension of the world, and in turn language, is still missing in many data curation efforts \cite{diaz2022crowdworksheets,scheuerman2021datasets}. A number of scholars have begun to not only document annotators' identity, but also develop principled approaches for obtaining a diversity of identities and perspectives in datasets.

\citet{aroyo2023dices} developed a dataset that specifically focuses on evaluating disagreement and diverse perspectives on conversational safety, and \cite{homan2023intersectionality} leverages this same dataset to demonstrate multilevel modeling as an approach for measuring annotation differences across a range of sociodemographic groups. Others have also successfully integrated annotator differences into model predictions, such as through personalized model tuning \cite{kumar2021designing}, and jury learning \cite{gordon2022jury}.

Disagreement among annotators in subjective tasks such as offensive language detection has roots beyond mere differences in socio-cultural backgrounds.
One such nuanced factor, often not studied in AI research, is morality. Moral considerations play significant roles in how humans navigate prejudicial thoughts and behaviors \cite{molina2016reflections}, often manifesting in language through offensive content. The interplay between morality and group identity \cite{reed2003moral} influences many aspects of our social dynamics, including perceptions, interactions, stereotypes, and prejudices. 
Moreover, research in computational social science addressing harmful language reveals a concurrent occurrence of moral sentiment alongside expressions of hatred directed at other social groups \cite{kennedy2023moral}.


Our data collection effort not only provides social factors and demographic information regarding annotators but also considers the moral values that may vary across regions and among individuals.
Such information facilitates drawing connections between annotations from culturally diverse annotators, the sociocultural norms shaping their environment, and the moral values they hold.

\section{\dataset Dataset}
\label{sec:data}

\begin{table}
\centering
    \scalebox{.75}{
    \vspace{-2cm}
    \begin{tabular}{lccccccccc}
    \toprule
        && \multicolumn{3}{c}{\textbf{Gender}} && \multicolumn{3}{c}{\textbf{Age}}\\\cmidrule{3-5}\cmidrule{7-9}
        \textbf{Region}                       & \multicolumn{1}{l}{\textbf{\#}} & \multicolumn{1}{l}{\textbf{M}} & \multicolumn{1}{l}{\textbf{W}} & \multicolumn{1}{l}{\textbf{Other}} && \multicolumn{1}{l}{\textbf{18--30}} & \multicolumn{1}{l}{\textbf{30--50}} & \multicolumn{1}{l}{\textbf{50+}}
         \\\midrule
        AC. &
        516 &
        306 &
        205 &
        5 &&
        269 &
        168 &
        79 \\
        ICS. &
        554 &
        308 &
        245 &
        1 &&
        237 &
        198 &
        119 \\
        LA. &
        549 &
        271 &
        275 &
        3 &&
        302 &
        176 &
        71 \\
        NA. &
        551 &
        220 &
        325 &
        6 &&
        263 &
        175 &
        113 \\
        Oc. &
        517 &
        203 &
        307 &
        7 &&
        161 &
        221 &
        135 \\
        Si. &
        540 &
        280 &
        249 &
        11 &&
        208 &
        228 &
        104 \\
        SSA. &
        530 &
        309 &
        219 &
        2 &&
        320 &
        157 &
        53 \\
        WE. &
        552 &
        252 &
        294 &
        6 &&
        259 &
        172 &
        121 \\
        \bottomrule                       
      \end{tabular}}
      \caption{Demographic distribution of annotators from each region, region names are shortened and represent: Arab Culture (AC.), Indian Cultural Sphere (ICS.), Latin America (LA.), North America (NA.), Oceania (Oc.), Sinosphere (Si.), Sun-Saharan Africa (SSA), and Western Europe (WE.).}
    \label{tab_stats}
\end{table}
 
In order to study a broad range of cultural perceptions of offensiveness, we recruited 4309 participants from 21 countries, representing eight geo-cultural regions, with each region represented by 2-4 countries (Table \ref{tab_stats}).\footnote{We based the categorization of regions loosely on the UN Sustainable Development Goals groupings \url{https://unstats.un.org/sdgs/indicators/regional-groups} with minor modifications: combining Australia, NZ and Oceania to ``Oceania'', and separating North America and Europe, to facilitate easier data collection.} We discuss the reasoning behind our selection of countries and regions in more depth in Appendix \ref{sec:regions}; however, the final selection of countries and regions was chosen to maximize cultural diversity while balancing participant access through our recruitment panel. 
Participants were recruited through an online survey pool, compensated in accordance to their local law, and were informed of the intended use of their responses. 
In order to capture the participants' perceptions of offensiveness, we asked each participant to 
annotate offensiveness of social media comments selected from Jigsaw datasets \cite{Jigsaw-toxic, Jigsaw-bias}. 
Furthermore, we also asked them to 
respond to a measurement of self-reported moral concerns, using the Moral Foundations Questionnaire \cite[MFQ-2;][]{graham2013moral,atari2022morality}.\footnote{The data card and dataset will be available upon the paper } 

\subsection{Recruitment}

Recruitment criteria account for various demographic attributes: (1) \textit{Region of residence}: we recruited at least 500 participants from each of the eight regions with at least 100 participants per country,
except for South Korea and Qatar where 
we managed to recruit only a smaller number of raters (See Table~\ref{tab:region-country}), 
(2) \textit{Gender}: within regions, we set a maximum limit of 60\% representations for Men and Women separately (for a loosely balanced representation of the two genders), while including options for selecting ``non-binary / third gender,'' ``prefer not to say,'' and ``prefer to self identify'' (with a textual input field). We recognize that collecting non-binary gender information is not safe for annotators in many countries, so we limited the specification of recruitment quota to binary genders to ensure consistency across countries.
(3) \textit{Age}: in each region at most 60\% of participants are 18 to 30 years old and at least 15\% are 50 years old or older. We specifically aimed to ensure adequate representation of annotators of age 50 or older, because this age group have lower engagement with crowdsourcing platforms but are equally impacted by technology advancements. 
Table \ref{tab_stats} provides the final distribution of participants across different demographic groups in each region.

We further set an exclusion criterion based on \textit{English fluency} since our study is done on English language text; we only selected participants who self-reported a high level of proficiency in reading and writing English.
We performed this study in the English language, as the most wide-spoken language across the globe, to simulate the most common data annotation settings, in which annotators (who are no necessarily English speakers) are asked to interact with and label textual data in English.
Additionally, we collected participants' self-reported subjective socio-economic status \cite{adler2000relationship} that may serve as a potential confound in follow-up analyses. 

\subsection{
Annotation items}
We performed this study in the English language. In order to collect textual items for participants to annotate, we selected items from Jigsaw's Toxic Comments Classification dataset \cite{Jigsaw-toxic}, and the Unintended Bias in Toxicity Classification dataset \cite{Jigsaw-bias}, both of which consist of social media comments 
labeled for toxicity.
We built a dataset of $N_{items}$ = 4554 consisting of three categories of items sampled from the above datasets: %
\paragraph{Random:} As the basic strategy, we randomly select 50\% of the data from items that are likely to evoke disagreement. 
To measure disagreements on each item, we averaged the toxicity scores assigned to the item in the original dataset, ranging from 0 (lowest toxicity) to 1 (highest toxicity). Items on the two ends of the range evoke no disagreement because all annotators labeled them either as toxic or non-toxic.
Therefore, we chose items with a normal distribution centered around a toxicity score of 0.5 (indicating highest disagreement) with a standard deviation of 0.2. 
\paragraph{Moral Sentiment: }Second, 10\% of the dataset consists of a balanced set of items include different moral sentiments, identified through a supervised moral language tagger trained on the MFTC dataset \cite{hoover2020moral}.
This strategy is aimed at enabling follow up studies to investigate potential content-level correlates of disagreements, particularly as previous computational social science studies on harmful language have shown specific correlation of moral sentiment with expressions of hatred \cite{kennedy2023moral}.
Our tagger identified very few items with moral sentiment throughout the dataset, selecting a balances set led to a set of 500 such items.
\paragraph{Social Group Mentions:} Finally, the rest (40\%) of the dataset consists of a balanced set of items that mention specific social group identities related to gender, sexual orientation, or religion (this information is provided in the Jigsaw's raw data). We specifically selected such items as online harmful language is largely directed at specific social groups and resonates real-world group conflicts.


\subsection{Annotation task}
Each participant was tasked with labeling 40 items on a 5-point Likert scale (from \textit{not offensive at all} to \textit{extremely offensive}). 
Half of the participants were provided with a note that defined \textit{extremely offensive language} as \textit{``profanity, strongly impolite, rude or vulgar language expressed with fighting or hurtful words in order to insult a targeted individual or group.''} Other participants were expected to label items based on their own definition of offensiveness. The latter group served as a control setting of participants who are expected to lean on their individual notion of offensiveness.\footnote{We did not explicitly ask participants to  provide their definition of offensiveness}.

In case of unfamiliarity with the annotation item, participants were asked to select the option \textit{``I do not understand this message.''}
Participants’ reliability was tested by 5 undeniably non-offensive, control questions randomly distributed among the 40-items annotation process. Those who failed at least one quality control check were removed, and not counted against our final set of 4309 participants (refer to Appendix \ref{sec:test} for test items). 
Each item in the final dataset was labeled by at least three participants from each region who passed the control check (a total of 24 labels). 
Participants were compensated at rates above the prevalent market rates for the task (which took at most 20 minutes, with a median of 13 minutes), and respecting the local regulations regarding minimum wage in their respective countries. 


\subsection{Moral Foundation Questionnaire}
After annotation, participants were also asked to fill out the Moral Foundations Questionnaire \cite[MFQ-2;][]{graham2013moral,atari2022morality}, which assesses their moral values along six different dimensions: \textit{Care}: ``avoiding emotional and physical damage to another individual,'' \textit{Equality}: ``equal treatment and equal outcome for individuals,'' \textit{Proportionality}: ``individuals getting rewarded in proportion to their merit or contribution,'' \textit{Authority}: ``deference toward legitimate authorities and the defense of traditions,'' \textit{Loyalty}: ``cooperating with ingroups and competing with outgroups,'' and \textit{Purity}: ``avoiding bodily and spiritual contamination and degradation'' \cite{atari2022morality}. We specifically rely on the MFQ-2 because it is developed and validated through extensive cross-cultural assessments of moral judgments. This characteristic makes the questionnaire a reliable tool for integrating a pluralistic definition of values into AI research.  
The questionnaire includes 36 statements to assess participants' priorities along each of the six foundations (see Figure \ref{fig:mfq} which shows one of the MFQ-2 questions in our survey). For instance, one MFQ-2 statement that targets the \textit{Care} foundation is: ``\textit{Everyone should try to comfort people who are going through something hard}''.
We aggregate each participant's responses to compute a value between 1 to 5 to capture their moral foundations along each of these dimensions.

\section{Analyses}
\begin{figure}[t]
    \centering
    \includegraphics[width=.23\textwidth]{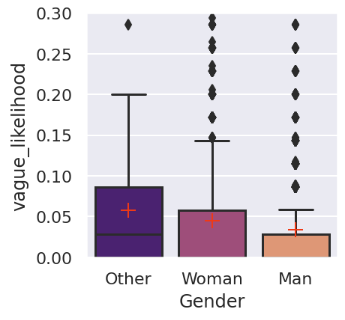}
    \includegraphics[width=.23\textwidth]{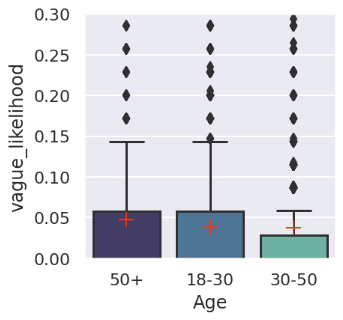}
    \includegraphics[width=.47\textwidth]{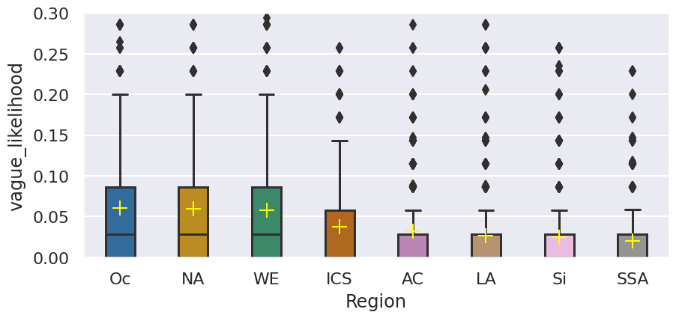}
    \caption{The likelihood of an annotator not understanding the message, grouped based on their socio-demographic information. Annotators identifying as Men, or of 50 years of old or younger are generally less likely to state they did not understand a message.}
    \label{fig:uncertainty-demo}
\end{figure}

Our analyses focus on annotators' varying perspectives and how shared social, cultural or moral attributes can help shed light on annotation behaviors. 
We begin by analyzing how different groups vary on expressing their lack of understanding the message by selecting the \textit{``I don't understand this message''} option.
We then study annotators' geo-cultural regions and moral values in relation to their annotations.
Specifically, we consider annotator clustering either based on their similar moral values or their region of residence, and assess in-group homogeneity  and out-group disagreements for clusters. The remainder of this section delves deeper into how groups of annotators from the same region or with similar moral values tend to label content differently. 

\subsection{Analysis of Lack Understanding}
We start our analyses by investigating the patterns of annotators not understanding the provided text.
While recent modeling efforts have shown the practical ways in which annotators' ambiguity or confidence can help inform the model. However, in many data annotation efforts, annotators' lack of understanding is either not captured or discarded. 
We ask whether specific groups of annotators are more likely to not understand the annotation item, and as a result, their responses are more likely to be discarded.  

We compared annotators with different demographics (along Gender, Age, and Region) on how likely they are to select the ``I don't understand'' answer (Figure \ref{fig:uncertainty-demo}). All further studies of the paper relies on the dataset after removing these answers. 

\paragraph{Gender:} When grouping annotators based on their gender or age, Men are overall less probable to state lack of understanding (M = .03, SD = .07), compared to Women (M = .05, SD = .08, $p$ < .001), and other genders (M = 0.06, SD = .07,  $p$ = .03). However, Women and other genders did not differently select this label ($p$ = .34). \paragraph{Age:} Participants who were aged 50 or more were distinctly more likely to state lack of understanding (M = .05, SD = .09), compared to 30 to 50 year-old (M = .04, SD = .08,  $p$ < .01), and 18 to 30 year-old participants (M = .04, SD = .07,  $p$ < .01). The difference of the latter two groups was insignificant ( $p$ = .85)
\paragraph{Region:} We further looked into the regional differences in not understanding the answers; a pairwise Tukey test shows that annotators from Oceania (M =  0.06, SD = 0.1), North America (M =  0.06, SD = 0.09), and Western Europe (M =  0.06, SD = 0.09) were all significantly more probably to state lack of understanding compared to Indian Cultural Sphere (M =  0.04, SD = 0.08), Arab Culture (M =  0.03, SD = 0.06), Latin America (M =  0.03, SD = 0.06), Sinosphere (M =  0.02, SD = 0.07), and Sub Saharan Africa (M =  0.02, SD = 0.05) with all $p$ values lower than .05. 

\begin{figure}[t!]
    \centering
    \begin{subfigure}[t]{0.47\textwidth}
        \centering
        \includegraphics[width=\textwidth]{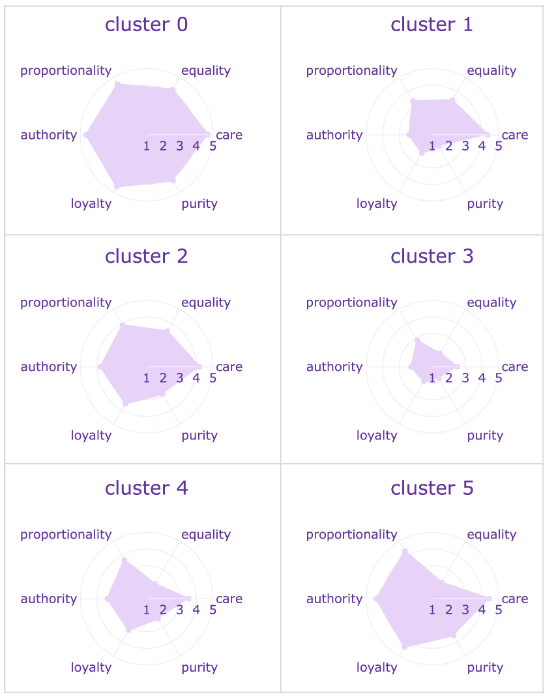}\vspace{2ex}
        \caption{The six moral clusters represented by the moral profile of their centroids. Clusters 0, 2 and 5 generally consist of participants who agreed more with the moral statements, with cluster 0 reporting the highest agreement. On the other hand, clusters 2, 3, and 4 report lower agreement with the moral statements, with cluster 3 consisting of participants who agreed the least. }
        \label{fig:moral-clusters}
    \end{subfigure}%
    \\\vspace{.1in}
    \begin{subfigure}[t]{0.47\textwidth}
        \centering
        \includegraphics[width=\textwidth]{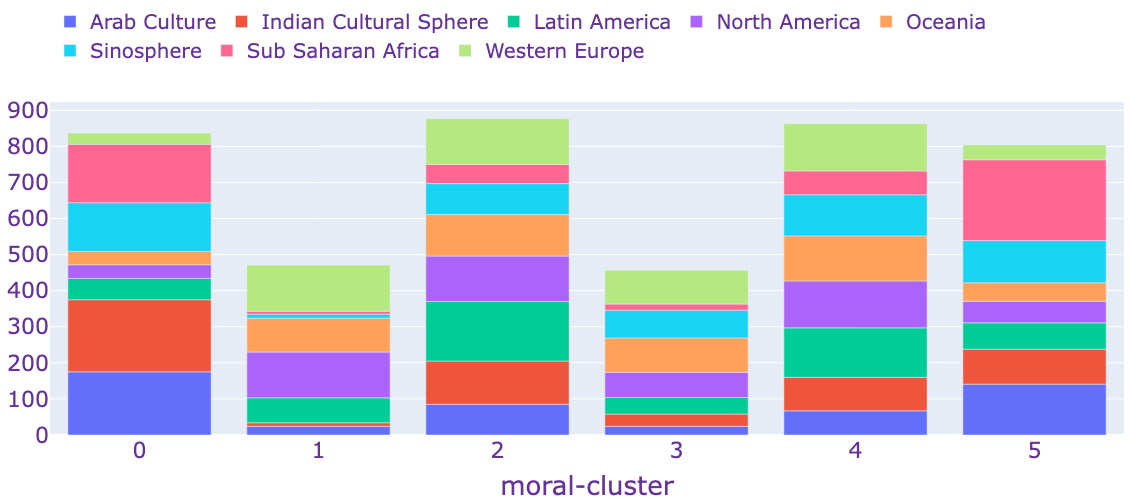}
        \caption{Distribution of participants from different regions across different moral clusters. Variances of regional presence are noticeable in several cases, e.g., cluster 0 mostly consists of participants from Indian Cultural Sphere, Arab Culture, and Sub-Saharan Africa.}
        \label{fig:moral-clusters-region}
    \end{subfigure}
\end{figure}

\subsection{Morally Aligned Annotators}

To systematically study annotators' perspectives with regard to varying moral values we first cluster annotators into groups with high internal moral similarity through a K-means algorithm, applying elbow method for finding the optimal number of clusters (see Appendix \ref{sec:kmeans}).
Figure \ref{fig:moral-clusters} represents the resulting six clusters by the average moral values of their members. Figure \ref{fig:moral-clusters-region} represents the distribution of annotators from different regions across the six moral clusters. As shown by the plots, regions have varying presence in the moral clusters;
cluster 0 consists of annotators who agreed most with all dimensions of the moral foundations questionnaire, most participants in this cluster are from Indian Cultural Sphere, Sub Saharan Africa and Arab Culture. 
On the other hand, cluster 3 includes annotators who agreed the least with MFQ-2 values along most dimensions;
while this cluster has the fewest annotators, most of them were from Western Europe, Oceania, and Sinosphere, in our data.
Other 4 clusters each have their specific distribution of moral values across the axes, that show the most prevalent moral values in the annotator pool.

\subsection{Disagreement among Groups}
Additionally, we explore the homogeneity of annotations within various clusters of annotators. We specifically compare moral clusters' homogeneity with the alternative clustering approach that considers annotators of the same region to have similar perceptions. We considered region as an alternative means for clustering annotators because collected annotations tend to vary significantly across regions and countries (the distribution of ratings collected from different countries is provided in Figure \ref{fig_stats}).
Inspired by \citet{prabhakaran2023framework}, we use the GAI metric which provides a measurement of perspective diversities within annotator groups. In other words, for each specific group of annotators, GAI provides the ratio of an in-group measurement of agreement to a cross-group measurement of cohesion. In our specific case, we measure in-group agreement through Inter-Rater Reliability \cite[IRR;][]{}, and cross-group cohesion through Cross-Replication Reliability \cite[XRR;][]{}. The GAI metric is then defined as the ratio to IRR to XRR, and value higher than 1 reports a group with blah blah. 


As Table \ref{tab:gai} shows, while the highest GAI score is achieved by one of the moral cluster (cluster 2, with low moral values on all axes), moral cluster in general have high variation in their homogeneity. On the other hand, regional clusters are generally more distinct in their perspectives.

\begin{table}[h!]

\centering
\scalebox{.8}{
\begin{tabular}{lllll}
\toprule
Dimension & Group &  IRR &  XRR & GAI \\
  \midrule
\multirow{ 8 }{*}{ Region }&  AC.
&\textbf{$\uparrow$0.13**}
&$\uparrow$0.11
&\textbf{$\uparrow$1.17*}
\\
&  ICS.
&$\downarrow$0.10
&\textbf{$\downarrow$0.10*}
&$\uparrow$1.04
\\
&  LA.
&\textbf{$\uparrow$0.13**}
&$\uparrow$0.11
&\textbf{$\uparrow$1.15*}
\\
&  NA.
&\textbf{$\uparrow$0.14**}
&$\uparrow$0.11
&\textbf{$\uparrow$1.31**}
\\
&  Oc.
&$\uparrow$0.12
&$\downarrow$0.10
&\textbf{$\uparrow$1.15*}
\\
&  Si.
&\textbf{$\downarrow$0.09*}
&\textbf{$\downarrow$0.09**}
&$\downarrow$1.00
\\
&  SSA.
&\textbf{$\uparrow$0.14**}
&$\downarrow$0.10
&\textbf{$\uparrow$1.36**}
\\
&  WE.
&\textbf{$\uparrow$0.14**}
&$\uparrow$0.11
&\textbf{$\uparrow$1.22**}
\\\midrule
&  0
&\textbf{$\uparrow$0.12*}
&\textbf{$\uparrow$0.12**}
&$\uparrow$1.05
\\
&  1
&$\uparrow$0.12
&$\uparrow$0.11
&$\uparrow$1.04
\\
Moral&  2
&\textbf{$\uparrow$0.18**}
&\textbf{$\uparrow$0.12**}
&\textbf{$\uparrow$1.46**}
\\
Cluster&  3
&\textbf{$\downarrow$0.07**}
&\textbf{$\downarrow$0.10**}
&\textbf{$\downarrow$0.75**}
\\
&  4
&$\uparrow$0.11
&$\uparrow$0.11
&$\uparrow$1.00
\\
&  5
&\textbf{$\downarrow$0.09*}
&\textbf{$\downarrow$0.09**}
&$\downarrow$0.97
\\
\bottomrule
 \end{tabular}}

\caption{Results for in-group and cross-group cohesion, and GAI. Significant results are in \textbf{bold}: * for significance at $p < 0.05$, ** for significance after Benjamini-Hochberg correction.
A $\downarrow$ (or $\uparrow$) means that the result is less (or greater) than expected under the null hypothesis. 
GAI results based on $C_X=$ XRR and $C_I=$ IRR. 
}
\label{tab:gai}

\end{table}

\subsection{Disagreement on Categories of Content}
\begin{figure}[t]
    \centering
    \begin{subfigure}[t]{0.47\textwidth}
        \centering
        \includegraphics[width=\textwidth]{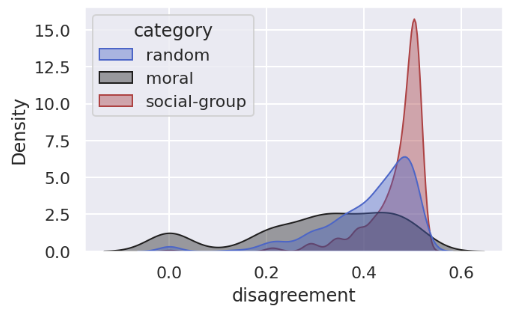}\\
        \caption{}
        \label{}
    \end{subfigure}
    \begin{subfigure}[t]{0.47\textwidth}
        \centering
        \includegraphics[width=\textwidth]{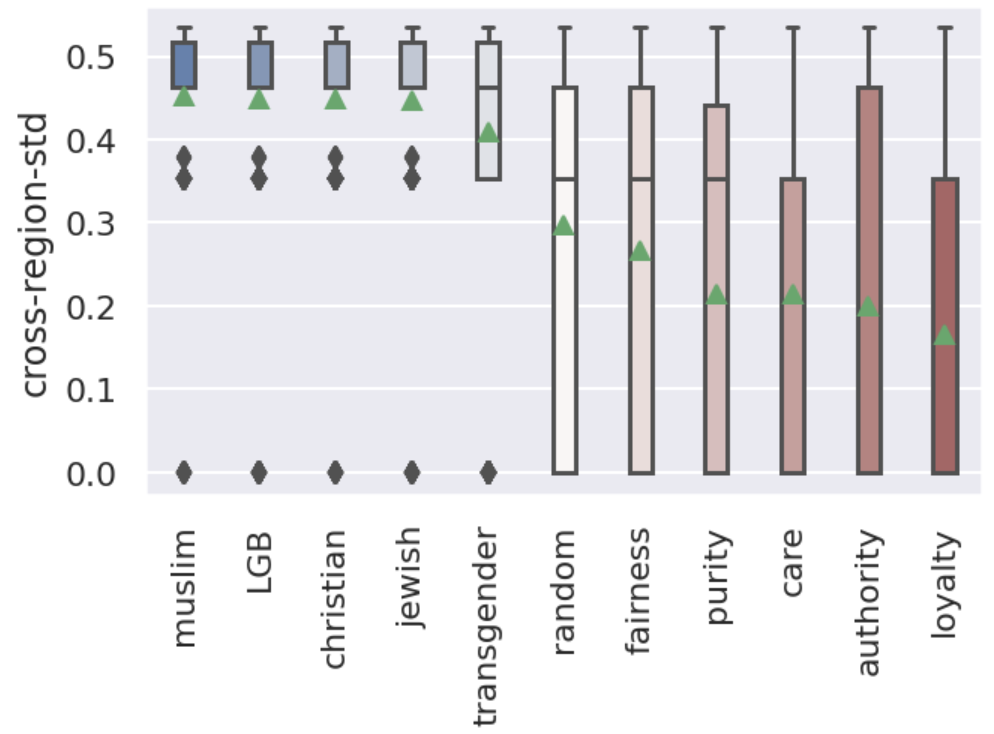}
        \caption{}
        \label{}
    \end{subfigure}
    \caption{Disagreement between regions on items from each category (a) and each sub-category (b). We considered the standard deviation of majority votes from different regions as the cross-regional disagreement. The plot shows that  items related to social groups (christian, transgender, jewish, muslim and LGB) generally evoke more disagreement compared to random items.}
    \label{fig:disagreement-cat}
\end{figure}

We further analyze the various types of content that annotators may label as offensive. As outlined in Section \ref{sec:data}, annotated items are chosen using three strategies: random selection, morality-based selection, and social identity-based selection. Figure \ref{fig:disagreement-cat} shows that annotators tend to have varying degrees of disagreement (calculated as the standard deviation of labels assigned to the item) when labeling items selected based on different strategies. As the plot shows, items that mention specific social identity groups evoke highest levels of disagreement (Mean = .47, SD = .06), significantly higher than items with moral sentiment (Mean = .31, SD = .16) and the randomly selected items (Mean = .41, SD = .10), both with $p$ < .001. It is important to note that our randomly selected items were deliberately chosen from those with high disagreement in the original 
Jigsaw dataset. Our analysis indicates that items mentioning social identity groups tend to evoke even more disagreement.

\begin{table}[]
    \centering
    \scalebox{.75}{
    \begin{tabular}{p{4.8cm}p{.9cm}p{.9cm}c}
    \toprule
        & \multicolumn{2}{c}{Offensive} & \\\cline{2-3}
        Item & Yes & No & Category  \\\midrule
        Transgender athletes..... Transgender students.....  This is what feminism gets you in society.  This was unheard of only a short 20 years ago. & AC, LA, NA, Oc & ICS, Si, SSA, WE & transgender \\\\
        Does pointing out that a growing majority of Americans support adultery (abortion, same sex marriage, even theft), change God's law an iota?&AC, ICS, LA, NA & Oc, Si, SSA, WE &LGB\\\\

        The women came out because he is a hypocrite, a child predator.  Come on trumpers,  defend the child predator Christian, lol & ICS, LA, Oc, SSA & AC, NA, Si, WE& christian\\
        \bottomrule
    \end{tabular}}
    \caption{Instances with highest disagreement across regions.}
    \label{tab:items-region}
\end{table}



In addition to disagreement between annotators, items can be labeled differently by various groups of annotators. In our case looking into the aggregated labels from each region demonstrates how recruiting annotators from specific regions could lead to having thoroughly different final dataset. Table \ref{tab:items-region} represents items with high cross-region disagreement.

\section{Discussion}
Research on safety considerations of large language models has mostly focused on evaluations of model harms through crowdsourced benchmarks \cite{srivastava2022beyond,wang2021adversarial}. However, while annotators from different regions are shown to have different perspectives regarding this task \cite{salminen2018online}, current benchmarks fail to represent the cultural and individual variations in human moral judgements about generated language and model outputs. They also lack comprehensive understanding of human values and cultural norms that drive diversity of perspectives in annotations. 
This work presents a cross-cultural experiment with participants across various cultural and demographic backgrounds. Our dataset captures valuable insights into human perceptions on offensive language, revealing demographic differences in annotation certainty, and regional, as well as moral psychological variations in perceiving offensiveness.


Our first analyses captures how participants with different demographic background might express their unfamiliarity with the annotation. In general, annotators not identifying as Men and annotators aged 50 and above are more likely to select the ``I don't understand'' option. Moreover, annotators from
Oceania, North America, and Western Europe were significantly more probably to state that they did not understand the message compared to Indian Cultural Sphere, Arab Culture, Latin America, Sinosphere, and Sub Saharan Africa.
Although we remove these responses for the remaining analyses and experiments in this paper, it is important to note this kind of uncertainty in annotating occurred disproportionately in these groups.

Our dataset also represent different categories of content within a well-known machine learning corpus, with annotators having varying levels of disagreement for labeling content from different categories. While items with moral sentiment are the least likely to evoke disagreement, items mentioning specific social groups are more likely to have a varying range of annotation. This finding replicates several previous findings on how group perception and stereotypes can affect harm perception targeting different social groups, in a cross-cultural context. Consequently, these findings underscore the need for further empirical research into social dynamics within diverse cultural contexts to better understand harmful language and mitigate harmful risks of language technologies for different social groups.

Furthermore, this study underscores the importance of incorporating cultural and individual perspectives into the development and evaluation of language models. By acknowledging and accounting for the diversity of moral judgments and values across different cultures and demographics, we can enhance the fairness and inclusivity of language technologies. This necessitates not only expanding the scope of data collection to include more diverse cultural perspectives but also implementing more nuanced evaluation metrics that consider the contextual nuances of language usage and interpretation. 
That paves a way towards
language models that are not only proficient in generating text but also sensitive to the diverse range of societal norms and values, ultimately fostering more respectful and inclusive interactions in digital spaces.

\section{Conclusion}

We introduce the \dataset dataset, which captures the results of a cross-cultural annotation experiment for understanding disagreements on perceiving offensiveness in language. Our findings reveal significant demographic and regional variations in perceptions of offensive language, underlining the necessity of incorporating diverse perspectives into reinforcement learning with human feedback. Additionally, the dataset showcases differences in annotation certainty and disagreement levels across various content categories, particularly concerning mentions of specific social groups. These findings underscore the imperative for further research into social dynamics within diverse cultural contexts to mitigate the risks associated with harmful language in language technologies and promote fairness and inclusivity in digital interactions.

\section*{Limitations}
In our work, we focus on moral foundations as a way to measure differences in values across groups; however, values can be measured in other ways, such as... Importantly, while our annotator sample represents diverse cultural perspectives, the items in our dataset are in English, which may explain the different rates of "I don't know" responses observed across regions. Moreover, English data likely features lower representation of certain content, such as offensive content about social groups, celebrations, or politics specific to certain regions and languages. In addition, to preserve our ability to compare data cross-culturally, we focused on demographic categories that are broadly recognized. As a result, we did not conduct analyses of demographic differences that are specific to particular cultural regions, such as caste, and we did not collect highly sensitive demographic information, such as sexual orientation. We acknowledge that salient social categories can differ greatly across geocultural reasons, therefore our selection of categories should not be considered exhaustive. Finally, our selection of countries within each cultural region was informed by access feasibility via our data collection platform, which may have introduced unexpected sampling biases.  

\section*{Ethics Statement}
In this work, we collected and modeled annotator responses primarily to demonstrate geocultural differences. Our results and approaches are not meant to be used to define user preferences or platform policies. For example, a subgroup's higher or lower tendency to identify content as offensive does not necessarily mean that content moderation policies should differ for that group. In addition, our work does not advocate for treating any particular cultural group's labels as more ``correct'' than those of another cultural group.

\section*{Acknowledgements}

\bibliography{bibs}

\begin{thebibliography}{48}
\expandafter\ifx\csname natexlab\endcsname\relax\def\natexlab#1{#1}\fi

\bibitem[{Adler et~al.(2000)Adler, Epel, Castellazzo, and
  Ickovics}]{adler2000relationship}
Nancy~E Adler, Elissa~S Epel, Grace Castellazzo, and Jeannette~R Ickovics.
  2000.
\newblock Relationship of subjective and objective social status with
  psychological and physiological functioning: Preliminary data in healthy,
  white women.
\newblock \emph{Health psychology}, 19(6):586.

\bibitem[{Aroyo et~al.(2023{\natexlab{a}})Aroyo, Diaz, Homan, Prabhakaran,
  Taylor, and Wang}]{aroyo2023reasonable}
Lora Aroyo, Mark Diaz, Christopher Homan, Vinodkumar Prabhakaran, Alex Taylor,
  and Ding Wang. 2023{\natexlab{a}}.
\newblock \href {http://arxiv.org/abs/2301.09406} {The reasonable effectiveness
  of diverse evaluation data}.

\bibitem[{Aroyo et~al.(2019)Aroyo, Dixon, Thain, Redfield, and
  Rosen}]{aroyo2019crowdsourcing}
Lora Aroyo, Lucas Dixon, Nithum Thain, Olivia Redfield, and Rachel Rosen. 2019.
\newblock \href
  {https://dl.acm.org/doi/pdf/10.1145/3308560.3317083?casa_token=KUbqXddeYqgAAAAA:oqSXNKuIXtMsCv4k5ZRAuBw9PdEANlmcWiGgQcR5_HL77kinmcWl9QmgpMsWhFPxa185gkJmEw5w}
  {Crowdsourcing subjective tasks: the case study of understanding toxicity in
  online discussions}.
\newblock In \emph{Companion proceedings of the 2019 world wide web
  conference}, pages 1100--1105.

\bibitem[{Aroyo et~al.(2023{\natexlab{b}})Aroyo, Taylor, D\'{i}az, Homan,
  Parrish, Serapio-Garc\'{i}a, Prabhakaran, and Wang}]{aroyo2023dices}
Lora Aroyo, Alex~S Taylor, Mark D\'{i}az, Christopher~M Homan, Alicia Parrish,
  Greg Serapio-Garc\'{i}a, Vinodkumar Prabhakaran, and Ding Wang.
  2023{\natexlab{b}}.
\newblock \href {https://arxiv.org/abs/2306.11247} {{DICES} dataset: Diversity
  in conversational ai evaluation for safety}.
\newblock In \emph{Proceedings of the Neural Information Processing Systems
  Track on Datasets and Benchmarks}.

\bibitem[{Aroyo and Welty(2015)}]{aroyo2015truth}
Lora Aroyo and Chris Welty. 2015.
\newblock Truth is a lie: Crowd truth and the seven myths of human annotation.
\newblock \emph{AI Magazine}, 36(1):15--24.

\bibitem[{Atari et~al.(2023)Atari, Haidt, Graham, Koleva, Stevens, and
  Dehghani}]{atari2022morality}
Mohammad Atari, Jonathan Haidt, Jesse Graham, Sena Koleva, Sean~T Stevens, and
  Morteza Dehghani. 2023.
\newblock Morality beyond the weird: How the nomological network of morality
  varies across cultures.
\newblock \emph{Journal of Personality and Social Psychology}, 125.

\bibitem[{Balkin(2017)}]{balkin2017digital}
Jack~M Balkin. 2017.
\newblock Digital speech and democratic culture: A theory of freedom of
  expression for the information society.
\newblock In \emph{Law and society approaches to cyberspace}, pages 325--382.
  Routledge.

\bibitem[{Basile et~al.(2021)Basile, Fell, Fornaciari, Hovy, Paun, Plank,
  Poesio, and Uma}]{basile2021need}
Valerio Basile, Michael Fell, Tommaso Fornaciari, Dirk Hovy, Silviu Paun,
  Barbara Plank, Massimo Poesio, and Alexandra Uma. 2021.
\newblock \href {https://doi.org/10.18653/v1/2021.bppf-1.3} {We need to
  consider disagreement in evaluation}.
\newblock In \emph{Proceedings of the 1st Workshop on Benchmarking: Past,
  Present and Future}, Online. Association for Computational Linguistics.

\bibitem[{Brannon(2019)}]{brannon2019free}
Valerie~C Brannon. 2019.
\newblock Free speech and the regulation of social media content.
\newblock \emph{Congressional Research Service}, 27.

\bibitem[{Davani et~al.(2023)Davani, Atari, Kennedy, and
  Dehghani}]{davani2023hate}
Aida~Mostafazadeh Davani, Mohammad Atari, Brendan Kennedy, and Morteza
  Dehghani. 2023.
\newblock \href
  {https://direct.mit.edu/tacl/article-pdf/doi/10.1162/tacl_a_00550/2075730/tacl_a_00550.pdf}
  {Hate speech classifiers learn normative social stereotypes}.
\newblock \emph{Transactions of the Association for Computational Linguistics},
  11:300--319.

\bibitem[{Davani et~al.(2022)Davani, D{\'\i}az, and
  Prabhakaran}]{davani2022dealing}
Aida~Mostafazadeh Davani, Mark D{\'\i}az, and Vinodkumar Prabhakaran. 2022.
\newblock \href {https://arxiv.org/pdf/2110.05719.pdf} {Dealing with
  disagreements: Looking beyond the majority vote in subjective annotations}.
\newblock \emph{Transactions of the Association for Computational Linguistics},
  10:92--110.

\bibitem[{D{\'\i}az et~al.(2022{\natexlab{a}})D{\'\i}az, Amironesei, Weidinger,
  and Gabriel}]{diaz2022accounting}
Mark D{\'\i}az, Razvan Amironesei, Laura Weidinger, and Iason Gabriel.
  2022{\natexlab{a}}.
\newblock Accounting for offensive speech as a practice of resistance.
\newblock In \emph{Proceedings of the sixth workshop on online abuse and harms
  (woah)}, pages 192--202.

\bibitem[{D{\'\i}az et~al.(2022{\natexlab{b}})D{\'\i}az, Kivlichan, Rosen,
  Baker, Amironesei, Prabhakaran, and Denton}]{diaz2022crowdworksheets}
Mark D{\'\i}az, Ian Kivlichan, Rachel Rosen, Dylan Baker, Razvan Amironesei,
  Vinodkumar Prabhakaran, and Emily Denton. 2022{\natexlab{b}}.
\newblock Crowdworksheets: Accounting for individual and collective identities
  underlying crowdsourced dataset annotation.
\newblock In \emph{2022 ACM Conference on Fairness, Accountability, and
  Transparency}, pages 2342--2351.

\bibitem[{Eickhoff(2018)}]{eickhoff2018cognitive}
Carsten Eickhoff. 2018.
\newblock \href
  {https://dl.acm.org/doi/pdf/10.1145/3159652.3159654?casa_token=f1VMBeUKBooAAAAA:XrttTdNDPXzDqGQIGyHyukTPBaVt2iopUYh839-jBLygMhOYprNMAR52Dp-mwzcXFJ8-3IojA5Xe}
  {Cognitive biases in crowdsourcing}.
\newblock In \emph{Proceedings of the eleventh ACM international conference on
  web search and data mining}, pages 162--170.

\bibitem[{Ferracane et~al.(2021)Ferracane, Durrett, Li, and
  Erk}]{ferracane-etal-2021-answer}
Elisa Ferracane, Greg Durrett, Junyi~Jessy Li, and Katrin Erk. 2021.
\newblock \href {https://doi.org/10.18653/v1/2021.naacl-main.129} {Did they
  answer? {S}ubjective acts and intents in conversational discourse}.
\newblock In \emph{Proceedings of the 2021 Conference of the North American
  Chapter of the Association for Computational Linguistics: Human Language
  Technologies}, pages 1626--1644, Online. Association for Computational
  Linguistics.

\bibitem[{Fornaciari et~al.(2021)Fornaciari, Uma, Paun, Plank, Hovy, and
  Poesio}]{fornaciari2021beyond}
Tommaso Fornaciari, Alexandra Uma, Silviu Paun, Barbara Plank, Dirk Hovy, and
  Massimo Poesio. 2021.
\newblock \href {https://aclanthology.org/2021.naacl-main.204.pdf} {Beyond
  black \& white: Leveraging annotator disagreement via soft-label multi-task
  learning}.
\newblock In \emph{Proceedings of the 2021 Conference of the North American
  Chapter of the Association for Computational Linguistics: Human Language
  Technologies}, pages 2591--2597.

\bibitem[{Founta et~al.(2018)Founta, Djouvas, Chatzakou, Leontiadis, Blackburn,
  Stringhini, Vakali, Sirivianos, and Kourtellis}]{founta2018large}
Antigoni Founta, Constantinos Djouvas, Despoina Chatzakou, Ilias Leontiadis,
  Jeremy Blackburn, Gianluca Stringhini, Athena Vakali, Michael Sirivianos, and
  Nicolas Kourtellis. 2018.
\newblock \href
  {https://ojs.aaai.org/index.php/ICWSM/article/download/14991/14841} {Large
  scale crowdsourcing and characterization of twitter abusive behavior}.
\newblock In \emph{Proceedings of the international AAAI conference on web and
  social media}, volume~12.

\bibitem[{Garten et~al.(2019)Garten, Kennedy, Hoover, Sagae, and
  Dehghani}]{garten2019incorporating}
Justin Garten, Brendan Kennedy, Joe Hoover, Kenji Sagae, and Morteza Dehghani.
  2019.
\newblock \href
  {https://onlinelibrary.wiley.com/doi/pdfdirect/10.1111/cogs.12701}
  {Incorporating demographic embeddings into language understanding}.
\newblock \emph{Cognitive science}, 43(1):e12701.

\bibitem[{Gordon et~al.(2022)Gordon, Lam, Park, Patel, Hancock, Hashimoto, and
  Bernstein}]{gordon2022jury}
Mitchell~L Gordon, Michelle~S Lam, Joon~Sung Park, Kayur Patel, Jeff Hancock,
  Tatsunori Hashimoto, and Michael~S Bernstein. 2022.
\newblock Jury learning: Integrating dissenting voices into machine learning
  models.
\newblock In \emph{Proceedings of the 2022 CHI Conference on Human Factors in
  Computing Systems}, pages 1--19.

\bibitem[{Graham et~al.(2013)Graham, Haidt, Koleva, Motyl, Iyer, Wojcik, and
  Ditto}]{graham2013moral}
Jesse Graham, Jonathan Haidt, Sena Koleva, Matt Motyl, Ravi Iyer, Sean~P
  Wojcik, and Peter~H Ditto. 2013.
\newblock \href
  {https://bpb-us-e2.wpmucdn.com/sites.uci.edu/dist/1/863/files/2020/06/Graham-et-al-2013.AESP_.pdf}
  {Moral foundations theory: The pragmatic validity of moral pluralism}.
\newblock In \emph{Advances in experimental social psychology}, volume~47,
  pages 55--130. Elsevier.

\bibitem[{Homan et~al.(2023)Homan, Serapio-Garc\'{i}a, Aroyo, D\'{i}az,
  Parrish, Prabhakaran, Taylor, and Wang}]{homan2023intersectionality}
Christopher~M Homan, Greg Serapio-Garc\'{i}a, Lora Aroyo, Mark D\'{i}az, Alicia
  Parrish, Vinodkumar Prabhakaran, Alex~S Taylor, and Ding Wang. 2023.
\newblock Intersectionality in conversational {AI} safety: How {B}ayesian
  multilevel models help understand diverse perceptions of safety.
\newblock \emph{arXiv preprint arXiv:2306.11530}.

\bibitem[{Hoover et~al.(2020)Hoover, Portillo-Wightman, Yeh, Havaldar, Davani,
  Lin, Kennedy, Atari, Kamel, Mendlen et~al.}]{hoover2020moral}
Joe Hoover, Gwenyth Portillo-Wightman, Leigh Yeh, Shreya Havaldar,
  Aida~Mostafazadeh Davani, Ying Lin, Brendan Kennedy, Mohammad Atari, Zahra
  Kamel, Madelyn Mendlen, et~al. 2020.
\newblock \href {https://journals.sagepub.com/doi/pdf/10.1177/1948550619876629}
  {Moral foundations twitter corpus: A collection of 35k tweets annotated for
  moral sentiment}.
\newblock \emph{Social Psychological and Personality Science},
  11(8):1057--1071.

\bibitem[{Hovy(2015)}]{hovy2015demographic}
Dirk Hovy. 2015.
\newblock \href {https://aclanthology.org/P15-1073.pdf} {Demographic factors
  improve classification performance}.
\newblock In \emph{Proceedings of the 53rd annual meeting of the Association
  for Computational Linguistics and the 7th international joint conference on
  natural language processing (volume 1: Long papers)}, pages 752--762.

\bibitem[{Hovy and Yang(2021)}]{hovy2021importance}
Dirk Hovy and Diyi Yang. 2021.
\newblock \href {https://aclanthology.org/2021.naacl-main.49.pdf} {The
  importance of modeling social factors of language: Theory and practice}.
\newblock In \emph{Proceedings of the 2021 Conference of the North American
  Chapter of the Association for Computational Linguistics: Human Language
  Technologies}, pages 588--602.

\bibitem[{Jigsaw(2018)}]{Jigsaw-toxic}
Jigsaw. 2018.
\newblock \href
  {https://www.kaggle.com/c/\\jigsaw-toxic-comment-classification-challenge/data}
  {Toxic comment classification challenge}.
\newblock Accessed: 2021-05-01.

\bibitem[{Jigsaw(2019)}]{Jigsaw-bias}
Jigsaw. 2019.
\newblock \href
  {https://www.kaggle.com/c/\\jigsaw-unintended-bias-in-toxicity-classification/data}
  {Unintended bias in toxicity classification}.
\newblock Accessed: 2021-05-01.

\bibitem[{Kennedy et~al.(2023)Kennedy, Golazizian, Trager, Atari, Hoover,
  Mostafazadeh~Davani, and Dehghani}]{kennedy2023moral}
Brendan Kennedy, Preni Golazizian, Jackson Trager, Mohammad Atari, Joe Hoover,
  Aida Mostafazadeh~Davani, and Morteza Dehghani. 2023.
\newblock The (moral) language of hate.
\newblock \emph{PNAS nexus}, 2(7):pgad210.

\bibitem[{Kennedy et~al.(2020)Kennedy, Bacon, Sahn, and von
  Vacano}]{kennedy2020constructing}
Chris~J Kennedy, Geoff Bacon, Alexander Sahn, and Claudia von Vacano. 2020.
\newblock \href {https://arxiv.org/pdf/2009.10277.pdf} {Constructing interval
  variables via faceted rasch measurement and multitask deep learning: a hate
  speech application}.
\newblock \emph{arXiv preprint arXiv:2009.10277}.

\bibitem[{Kiritchenko et~al.(2021)Kiritchenko, Nejadgholi, and
  Fraser}]{kiritchenko2021confronting}
Svetlana Kiritchenko, Isar Nejadgholi, and Kathleen~C Fraser. 2021.
\newblock Confronting abusive language online: A survey from the ethical and
  human rights perspective.
\newblock \emph{Journal of Artificial Intelligence Research}, 71:431--478.

\bibitem[{Kumar et~al.(2021)Kumar, Kelley, Consolvo, Mason, Bursztein,
  Durumeric, Thomas, and Bailey}]{kumar2021designing}
Deepak Kumar, Patrick~Gage Kelley, Sunny Consolvo, Joshua Mason, Elie
  Bursztein, Zakir Durumeric, Kurt Thomas, and Michael Bailey. 2021.
\newblock Designing toxic content classification for a diversity of
  perspectives.
\newblock In \emph{Seventeenth Symposium on Usable Privacy and Security (SOUPS
  2021)}, pages 299--318.

\bibitem[{Molina et~al.(2016)Molina, Tropp, and Goode}]{molina2016reflections}
Ludwin~E Molina, Linda~R Tropp, and Chris Goode. 2016.
\newblock Reflections on prejudice and intergroup relations.
\newblock \emph{Current Opinion in Psychology}, 11:120--124.

\bibitem[{Orlikowski et~al.(2023)Orlikowski, R{\"o}ttger, Cimiano, and
  Hovy}]{orlikowski-etal-2023-ecological}
Matthias Orlikowski, Paul R{\"o}ttger, Philipp Cimiano, and Dirk Hovy. 2023.
\newblock \href {https://doi.org/10.18653/v1/2023.acl-short.88} {The ecological
  fallacy in annotation: Modeling human label variation goes beyond
  sociodemographics}.
\newblock In \emph{Proceedings of the 61st Annual Meeting of the Association
  for Computational Linguistics (Volume 2: Short Papers)}, pages 1017--1029,
  Toronto, Canada. Association for Computational Linguistics.

\bibitem[{Parrish et~al.(2023)Parrish, Laszlo, and Aroyo}]{parrish2023picture}
Alicia Parrish, Sarah Laszlo, and Lora Aroyo. 2023.
\newblock " is a picture of a bird a bird": Policy recommendations for dealing
  with ambiguity in machine vision models.
\newblock \emph{arXiv preprint arXiv:2306.15777}.

\bibitem[{Pavlick and Kwiatkowski(2019)}]{pavlick2019inherent}
Ellie Pavlick and Tom Kwiatkowski. 2019.
\newblock Inherent disagreements in human textual inferences.
\newblock \emph{Transactions of the Association for Computational Linguistics},
  7:677--694.

\bibitem[{Plank et~al.(2014)Plank, Hovy, and S{\o}gaard}]{plank2014learning}
Barbara Plank, Dirk Hovy, and Anders S{\o}gaard. 2014.
\newblock \href {https://aclanthology.org/E14-1078.pdf} {Learning
  part-of-speech taggers with inter-annotator agreement loss}.
\newblock In \emph{Proceedings of the 14th Conference of the European Chapter
  of the Association for Computational Linguistics}, pages 742--751.

\bibitem[{Prabhakaran et~al.(2023)Prabhakaran, Homan, Aroyo, Parrish, Taylor,
  D{\'\i}az, and Wang}]{prabhakaran2023framework}
Vinodkumar Prabhakaran, Christopher Homan, Lora Aroyo, Alicia Parrish, Alex
  Taylor, Mark D{\'\i}az, and Ding Wang. 2023.
\newblock A framework to assess (dis) agreement among diverse rater groups.
\newblock \emph{arXiv preprint arXiv:2311.05074}.

\bibitem[{Prabhakaran et~al.(2021)Prabhakaran, Mostafazadeh~Davani, and
  Diaz}]{prabhakaran2021releasing}
Vinodkumar Prabhakaran, Aida Mostafazadeh~Davani, and Mark Diaz. 2021.
\newblock \href {https://doi.org/10.18653/v1/2021.law-1.14} {On releasing
  annotator-level labels and information in datasets}.
\newblock In \emph{Proceedings of The Joint 15th Linguistic Annotation Workshop
  (LAW) and 3rd Designing Meaning Representations (DMR) Workshop}, pages
  133--138, Punta Cana, Dominican Republic. Association for Computational
  Linguistics.

\bibitem[{Reed~II and Aquino(2003)}]{reed2003moral}
Americus Reed~II and Karl~F Aquino. 2003.
\newblock Moral identity and the expanding circle of moral regard toward
  out-groups.
\newblock \emph{Journal of personality and social psychology}, 84(6):1270.

\bibitem[{Rottger et~al.(2022)Rottger, Vidgen, Hovy, and
  Pierrehumbert}]{rottger2021two}
Paul Rottger, Bertie Vidgen, Dirk Hovy, and Janet Pierrehumbert. 2022.
\newblock \href {https://doi.org/10.18653/v1/2022.naacl-main.13} {Two
  contrasting data annotation paradigms for subjective {NLP} tasks}.
\newblock In \emph{Proceedings of the 2022 Conference of the North American
  Chapter of the Association for Computational Linguistics: Human Language
  Technologies}, pages 175--190, Seattle, United States. Association for
  Computational Linguistics.

\bibitem[{Salminen et~al.(2019)Salminen, Almerekhi, Kamel, Jung, and
  Jansen}]{salminen2019online}
Joni Salminen, Hind Almerekhi, Ahmed~Mohamed Kamel, Soon-gyo Jung, and
  Bernard~J Jansen. 2019.
\newblock \href {https://dl.acm.org/doi/pdf/10.1145/3295750.3298954} {Online
  hate ratings vary by extremes: A statistical analysis}.
\newblock In \emph{Proceedings of the 2019 Conference on Human Information
  Interaction and Retrieval}, pages 213--217.

\bibitem[{Salminen et~al.(2018)Salminen, Veronesi, Almerekhi, Jung, and
  Jansen}]{salminen2018online}
Joni Salminen, Fabio Veronesi, Hind Almerekhi, Soon-Gvo Jung, and Bernard~J
  Jansen. 2018.
\newblock \href
  {http://www.bernardjjansen.com/uploads/2/4/1/8/24188166/jansen_onlinehate2018.pdf}
  {Online hate interpretation varies by country, but more by individual: A
  statistical analysis using crowdsourced ratings}.
\newblock In \emph{2018 fifth international conference on social networks
  analysis, management and security (snams)}, pages 88--94. IEEE.

\bibitem[{Scheuerman et~al.(2021)Scheuerman, Hanna, and
  Denton}]{scheuerman2021datasets}
Morgan~Klaus Scheuerman, Alex Hanna, and Emily Denton. 2021.
\newblock Do datasets have politics? {D}isciplinary values in computer vision
  dataset development.
\newblock \emph{Proceedings of the ACM on Human-Computer Interaction},
  5(CSCW2):1--37.

\bibitem[{Srivastava et~al.(2023)Srivastava, Rastogi, Rao, Shoeb, Abid, Fisch,
  Brown, Santoro, Gupta, Garriga-Alonso, Kluska, Lewkowycz, Agarwal, Power,
  Ray, Warstadt, Kocurek, Safaya, Tazarv, Xiang, Parrish, Nie, Hussain, Askell,
  Dsouza, Slone, Rahane, Iyer, Andreassen, Madotto, Santilli, Stuhlmüller,
  Dai, La, Lampinen, Zou, Jiang, Chen, Vuong, Gupta, Gottardi, Norelli,
  Venkatesh, Gholamidavoodi, Tabassum, Menezes, Kirubarajan, Mullokandov,
  Sabharwal, Herrick, Efrat, Erdem, Karakaş, Roberts, Loe, Zoph, Bojanowski,
  Özyurt, Hedayatnia, Neyshabur, Inden, Stein, Ekmekci, Lin, Howald, Orinion,
  Diao, Dour, Stinson, Argueta, Ramírez, Singh, Rathkopf, Meng, Baral, Wu,
  Callison-Burch, Waites, Voigt, Manning, Potts, Ramirez, Rivera, Siro, Raffel,
  Ashcraft, Garbacea, Sileo, Garrette, Hendrycks, Kilman, Roth, Freeman,
  Khashabi, Levy, González, Perszyk, Hernandez, Chen, Ippolito, Gilboa, Dohan,
  Drakard, Jurgens, Datta, Ganguli, Emelin, Kleyko, Yuret, Chen, Tam, Hupkes,
  Misra, Buzan, Mollo, Yang, Lee, Schrader, Shutova, Cubuk, Segal, Hagerman,
  Barnes, Donoway, Pavlick, Rodola, Lam, Chu, Tang, Erdem, Chang, Chi, Dyer,
  Jerzak, Kim, Manyasi, Zheltonozhskii, Xia, Siar, Martínez-Plumed, Happé,
  Chollet, Rong, Mishra, Winata, de~Melo, Kruszewski, Parascandolo, Mariani,
  Wang, Jaimovitch-López, Betz, Gur-Ari, Galijasevic, Kim, Rashkin,
  Hajishirzi, Mehta, Bogar, Shevlin, Schütze, Yakura, Zhang, Wong, Ng, Noble,
  Jumelet, Geissinger, Kernion, Hilton, Lee, Fisac, Simon, Koppel, Zheng, Zou,
  Kocoń, Thompson, Wingfield, Kaplan, Radom, Sohl-Dickstein, Phang, Wei,
  Yosinski, Novikova, Bosscher, Marsh, Kim, Taal, Engel, Alabi, Xu, Song, Tang,
  Waweru, Burden, Miller, Balis, Batchelder, Berant, Frohberg, Rozen,
  Hernandez-Orallo, Boudeman, Guerr, Jones, Tenenbaum, Rule, Chua, Kanclerz,
  Livescu, Krauth, Gopalakrishnan, Ignatyeva, Markert, Dhole, Gimpel, Omondi,
  Mathewson, Chiafullo, Shkaruta, Shridhar, McDonell, Richardson, Reynolds,
  Gao, Zhang, Dugan, Qin, Contreras-Ochando, Morency, Moschella, Lam, Noble,
  Schmidt, He, Colón, Metz, Şenel, Bosma, Sap, ter Hoeve, Farooqi, Faruqui,
  Mazeika, Baturan, Marelli, Maru, Quintana, Tolkiehn, Giulianelli, Lewis,
  Potthast, Leavitt, Hagen, Schubert, Baitemirova, Arnaud, McElrath, Yee,
  Cohen, Gu, Ivanitskiy, Starritt, Strube, Swędrowski, Bevilacqua, Yasunaga,
  Kale, Cain, Xu, Suzgun, Walker, Tiwari, Bansal, Aminnaseri, Geva, Gheini, T,
  Peng, Chi, Lee, Krakover, Cameron, Roberts, Doiron, Martinez, Nangia,
  Deckers, Muennighoff, Keskar, Iyer, Constant, Fiedel, Wen, Zhang, Agha,
  Elbaghdadi, Levy, Evans, Casares, Doshi, Fung, Liang, Vicol,
  Alipoormolabashi, Liao, Liang, Chang, Eckersley, Htut, Hwang, Miłkowski,
  Patil, Pezeshkpour, Oli, Mei, Lyu, Chen, Banjade, Rudolph, Gabriel, Habacker,
  Risco, Millière, Garg, Barnes, Saurous, Arakawa, Raymaekers, Frank, Sikand,
  Novak, Sitelew, LeBras, Liu, Jacobs, Zhang, Salakhutdinov, Chi, Lee, Stovall,
  Teehan, Yang, Singh, Mohammad, Anand, Dillavou, Shleifer, Wiseman, Gruetter,
  Bowman, Schoenholz, Han, Kwatra, Rous, Ghazarian, Ghosh, Casey, Bischoff,
  Gehrmann, Schuster, Sadeghi, Hamdan, Zhou, Srivastava, Shi, Singh, Asaadi,
  Gu, Pachchigar, Toshniwal, Upadhyay, Shyamolima, Debnath, Shakeri, Thormeyer,
  Melzi, Reddy, Makini, Lee, Torene, Hatwar, Dehaene, Divic, Ermon, Biderman,
  Lin, Prasad, Piantadosi, Shieber, Misherghi, Kiritchenko, Mishra, Linzen,
  Schuster, Li, Yu, Ali, Hashimoto, Wu, Desbordes, Rothschild, Phan, Wang,
  Nkinyili, Schick, Kornev, Tunduny, Gerstenberg, Chang, Neeraj, Khot, Shultz,
  Shaham, Misra, Demberg, Nyamai, Raunak, Ramasesh, Prabhu, Padmakumar,
  Srikumar, Fedus, Saunders, Zhang, Vossen, Ren, Tong, Zhao, Wu, Shen,
  Yaghoobzadeh, Lakretz, Song, Bahri, Choi, Yang, Hao, Chen, Belinkov, Hou,
  Hou, Bai, Seid, Zhao, Wang, Wang, Wang, and Wu}]{srivastava2022beyond}
Aarohi Srivastava, Abhinav Rastogi, Abhishek Rao, Abu Awal~Md Shoeb, Abubakar
  Abid, Adam Fisch, Adam~R. Brown, Adam Santoro, Aditya Gupta, Adrià
  Garriga-Alonso, Agnieszka Kluska, Aitor Lewkowycz, Akshat Agarwal, Alethea
  Power, Alex Ray, Alex Warstadt, Alexander~W. Kocurek, Ali Safaya, Ali Tazarv,
  Alice Xiang, Alicia Parrish, Allen Nie, Aman Hussain, Amanda Askell, Amanda
  Dsouza, Ambrose Slone, Ameet Rahane, Anantharaman~S. Iyer, Anders Andreassen,
  Andrea Madotto, Andrea Santilli, Andreas Stuhlmüller, Andrew Dai, Andrew La,
  Andrew Lampinen, Andy Zou, Angela Jiang, Angelica Chen, Anh Vuong, Animesh
  Gupta, Anna Gottardi, Antonio Norelli, Anu Venkatesh, Arash Gholamidavoodi,
  Arfa Tabassum, Arul Menezes, Arun Kirubarajan, Asher Mullokandov, Ashish
  Sabharwal, Austin Herrick, Avia Efrat, Aykut Erdem, Ayla Karakaş, B.~Ryan
  Roberts, Bao~Sheng Loe, Barret Zoph, Bartłomiej Bojanowski, Batuhan Özyurt,
  Behnam Hedayatnia, Behnam Neyshabur, Benjamin Inden, Benno Stein, Berk
  Ekmekci, Bill~Yuchen Lin, Blake Howald, Bryan Orinion, Cameron Diao, Cameron
  Dour, Catherine Stinson, Cedrick Argueta, César~Ferri Ramírez, Chandan
  Singh, Charles Rathkopf, Chenlin Meng, Chitta Baral, Chiyu Wu, Chris
  Callison-Burch, Chris Waites, Christian Voigt, Christopher~D. Manning,
  Christopher Potts, Cindy Ramirez, Clara~E. Rivera, Clemencia Siro, Colin
  Raffel, Courtney Ashcraft, Cristina Garbacea, Damien Sileo, Dan Garrette, Dan
  Hendrycks, Dan Kilman, Dan Roth, Daniel Freeman, Daniel Khashabi, Daniel
  Levy, Daniel~Moseguí González, Danielle Perszyk, Danny Hernandez, Danqi
  Chen, Daphne Ippolito, Dar Gilboa, David Dohan, David Drakard, David Jurgens,
  Debajyoti Datta, Deep Ganguli, Denis Emelin, Denis Kleyko, Deniz Yuret, Derek
  Chen, Derek Tam, Dieuwke Hupkes, Diganta Misra, Dilyar Buzan, Dimitri~Coelho
  Mollo, Diyi Yang, Dong-Ho Lee, Dylan Schrader, Ekaterina Shutova, Ekin~Dogus
  Cubuk, Elad Segal, Eleanor Hagerman, Elizabeth Barnes, Elizabeth Donoway,
  Ellie Pavlick, Emanuele Rodola, Emma Lam, Eric Chu, Eric Tang, Erkut Erdem,
  Ernie Chang, Ethan~A. Chi, Ethan Dyer, Ethan Jerzak, Ethan Kim, Eunice~Engefu
  Manyasi, Evgenii Zheltonozhskii, Fanyue Xia, Fatemeh Siar, Fernando
  Martínez-Plumed, Francesca Happé, Francois Chollet, Frieda Rong, Gaurav
  Mishra, Genta~Indra Winata, Gerard de~Melo, Germán Kruszewski, Giambattista
  Parascandolo, Giorgio Mariani, Gloria Wang, Gonzalo Jaimovitch-López, Gregor
  Betz, Guy Gur-Ari, Hana Galijasevic, Hannah Kim, Hannah Rashkin, Hannaneh
  Hajishirzi, Harsh Mehta, Hayden Bogar, Henry Shevlin, Hinrich Schütze,
  Hiromu Yakura, Hongming Zhang, Hugh~Mee Wong, Ian Ng, Isaac Noble, Jaap
  Jumelet, Jack Geissinger, Jackson Kernion, Jacob Hilton, Jaehoon Lee,
  Jaime~Fernández Fisac, James~B. Simon, James Koppel, James Zheng, James Zou,
  Jan Kocoń, Jana Thompson, Janelle Wingfield, Jared Kaplan, Jarema Radom,
  Jascha Sohl-Dickstein, Jason Phang, Jason Wei, Jason Yosinski, Jekaterina
  Novikova, Jelle Bosscher, Jennifer Marsh, Jeremy Kim, Jeroen Taal, Jesse
  Engel, Jesujoba Alabi, Jiacheng Xu, Jiaming Song, Jillian Tang, Joan Waweru,
  John Burden, John Miller, John~U. Balis, Jonathan Batchelder, Jonathan
  Berant, Jörg Frohberg, Jos Rozen, Jose Hernandez-Orallo, Joseph Boudeman,
  Joseph Guerr, Joseph Jones, Joshua~B. Tenenbaum, Joshua~S. Rule, Joyce Chua,
  Kamil Kanclerz, Karen Livescu, Karl Krauth, Karthik Gopalakrishnan, Katerina
  Ignatyeva, Katja Markert, Kaustubh~D. Dhole, Kevin Gimpel, Kevin Omondi, Kory
  Mathewson, Kristen Chiafullo, Ksenia Shkaruta, Kumar Shridhar, Kyle McDonell,
  Kyle Richardson, Laria Reynolds, Leo Gao, Li~Zhang, Liam Dugan, Lianhui Qin,
  Lidia Contreras-Ochando, Louis-Philippe Morency, Luca Moschella, Lucas Lam,
  Lucy Noble, Ludwig Schmidt, Luheng He, Luis~Oliveros Colón, Luke Metz,
  Lütfi~Kerem Şenel, Maarten Bosma, Maarten Sap, Maartje ter Hoeve, Maheen
  Farooqi, Manaal Faruqui, Mantas Mazeika, Marco Baturan, Marco Marelli, Marco
  Maru, Maria Jose~Ramírez Quintana, Marie Tolkiehn, Mario Giulianelli, Martha
  Lewis, Martin Potthast, Matthew~L. Leavitt, Matthias Hagen, Mátyás
  Schubert, Medina~Orduna Baitemirova, Melody Arnaud, Melvin McElrath,
  Michael~A. Yee, Michael Cohen, Michael Gu, Michael Ivanitskiy, Michael
  Starritt, Michael Strube, Michał Swędrowski, Michele Bevilacqua, Michihiro
  Yasunaga, Mihir Kale, Mike Cain, Mimee Xu, Mirac Suzgun, Mitch Walker,
  Mo~Tiwari, Mohit Bansal, Moin Aminnaseri, Mor Geva, Mozhdeh Gheini,
  Mukund~Varma T, Nanyun Peng, Nathan~A. Chi, Nayeon Lee, Neta Gur-Ari
  Krakover, Nicholas Cameron, Nicholas Roberts, Nick Doiron, Nicole Martinez,
  Nikita Nangia, Niklas Deckers, Niklas Muennighoff, Nitish~Shirish Keskar,
  Niveditha~S. Iyer, Noah Constant, Noah Fiedel, Nuan Wen, Oliver Zhang, Omar
  Agha, Omar Elbaghdadi, Omer Levy, Owain Evans, Pablo Antonio~Moreno Casares,
  Parth Doshi, Pascale Fung, Paul~Pu Liang, Paul Vicol, Pegah Alipoormolabashi,
  Peiyuan Liao, Percy Liang, Peter Chang, Peter Eckersley, Phu~Mon Htut, Pinyu
  Hwang, Piotr Miłkowski, Piyush Patil, Pouya Pezeshkpour, Priti Oli, Qiaozhu
  Mei, Qing Lyu, Qinlang Chen, Rabin Banjade, Rachel~Etta Rudolph, Raefer
  Gabriel, Rahel Habacker, Ramon Risco, Raphaël Millière, Rhythm Garg,
  Richard Barnes, Rif~A. Saurous, Riku Arakawa, Robbe Raymaekers, Robert Frank,
  Rohan Sikand, Roman Novak, Roman Sitelew, Ronan LeBras, Rosanne Liu, Rowan
  Jacobs, Rui Zhang, Ruslan Salakhutdinov, Ryan Chi, Ryan Lee, Ryan Stovall,
  Ryan Teehan, Rylan Yang, Sahib Singh, Saif~M. Mohammad, Sajant Anand, Sam
  Dillavou, Sam Shleifer, Sam Wiseman, Samuel Gruetter, Samuel~R. Bowman,
  Samuel~S. Schoenholz, Sanghyun Han, Sanjeev Kwatra, Sarah~A. Rous, Sarik
  Ghazarian, Sayan Ghosh, Sean Casey, Sebastian Bischoff, Sebastian Gehrmann,
  Sebastian Schuster, Sepideh Sadeghi, Shadi Hamdan, Sharon Zhou, Shashank
  Srivastava, Sherry Shi, Shikhar Singh, Shima Asaadi, Shixiang~Shane Gu, Shubh
  Pachchigar, Shubham Toshniwal, Shyam Upadhyay, Shyamolima, Debnath, Siamak
  Shakeri, Simon Thormeyer, Simone Melzi, Siva Reddy, Sneha~Priscilla Makini,
  Soo-Hwan Lee, Spencer Torene, Sriharsha Hatwar, Stanislas Dehaene, Stefan
  Divic, Stefano Ermon, Stella Biderman, Stephanie Lin, Stephen Prasad,
  Steven~T. Piantadosi, Stuart~M. Shieber, Summer Misherghi, Svetlana
  Kiritchenko, Swaroop Mishra, Tal Linzen, Tal Schuster, Tao Li, Tao Yu, Tariq
  Ali, Tatsu Hashimoto, Te-Lin Wu, Théo Desbordes, Theodore Rothschild, Thomas
  Phan, Tianle Wang, Tiberius Nkinyili, Timo Schick, Timofei Kornev, Titus
  Tunduny, Tobias Gerstenberg, Trenton Chang, Trishala Neeraj, Tushar Khot,
  Tyler Shultz, Uri Shaham, Vedant Misra, Vera Demberg, Victoria Nyamai, Vikas
  Raunak, Vinay Ramasesh, Vinay~Uday Prabhu, Vishakh Padmakumar, Vivek
  Srikumar, William Fedus, William Saunders, William Zhang, Wout Vossen, Xiang
  Ren, Xiaoyu Tong, Xinran Zhao, Xinyi Wu, Xudong Shen, Yadollah Yaghoobzadeh,
  Yair Lakretz, Yangqiu Song, Yasaman Bahri, Yejin Choi, Yichi Yang, Yiding
  Hao, Yifu Chen, Yonatan Belinkov, Yu~Hou, Yufang Hou, Yuntao Bai, Zachary
  Seid, Zhuoye Zhao, Zijian Wang, Zijie~J. Wang, Zirui Wang, and Ziyi Wu. 2023.
\newblock \href {http://arxiv.org/abs/2206.04615} {Beyond the imitation game:
  Quantifying and extrapolating the capabilities of language models}.

\bibitem[{Uma et~al.(2021)Uma, Fornaciari, Hovy, Paun, Plank, and
  Poesio}]{uma2021learning}
Alexandra~N Uma, Tommaso Fornaciari, Dirk Hovy, Silviu Paun, Barbara Plank, and
  Massimo Poesio. 2021.
\newblock \href
  {https://www.jair.org/index.php/jair/article/download/12752/26751} {Learning
  from disagreement: A survey}.
\newblock \emph{Journal of Artificial Intelligence Research}, 72:1385--1470.

\bibitem[{Wang et~al.(2022)Wang, Xu, Wang, Gan, Cheng, Gao, Awadallah, and
  Li}]{wang2021adversarial}
Boxin Wang, Chejian Xu, Shuohang Wang, Zhe Gan, Yu~Cheng, Jianfeng Gao,
  Ahmed~Hassan Awadallah, and Bo~Li. 2022.
\newblock \href {http://arxiv.org/abs/2111.02840} {Adversarial {GLUE}: A
  multi-task benchmark for robustness evaluation of language models}.

\bibitem[{Waseem(2016)}]{talat2016you}
Zeerak Waseem. 2016.
\newblock \href {https://aclanthology.org/W16-5618.pdf} {Are you a racist or am
  {I} seeing things? {A}nnotator influence on hate speech detection on
  twitter}.
\newblock In \emph{Proceedings of the first workshop on NLP and computational
  social science}, pages 138--142.

\bibitem[{Waseem et~al.(2021)Waseem, Lulz, Bingel, and
  Augenstein}]{talat2021disembodied}
Zeerak Waseem, Smarika Lulz, Joachim Bingel, and Isabelle Augenstein. 2021.
\newblock \href {http://arxiv.org/abs/2101.11974} {Disembodied machine
  learning: On the illusion of objectivity in nlp}.

\bibitem[{Wulczyn et~al.(2017)Wulczyn, Thain, and Dixon}]{wulczyn2017ex}
Ellery Wulczyn, Nithum Thain, and Lucas Dixon. 2017.
\newblock \href
  {https://arxiv.org/pdf/1610.08914.pdf?_gclid=5aec59ba53a138.82841565-5aec59ba53a189.59055081&_utm_source=xakep&_utm_campaign=mention114889&_utm_medium=inline&_utm_content=lnk530117377130}
  {Ex machina: Personal attacks seen at scale}.
\newblock In \emph{Proceedings of the 26th international conference on world
  wide web}, pages 1391--1399.

\end{thebibliography}
\bibliographystyle{acl_natbib}

\appendix

\section{Appendix}
\label{sec:appendix}
\begin{table}[ht]
    \centering
    \scalebox{.9}{
    \begin{tabular}{m{4em} m{10em} m{4em}}
    \toprule
        Foundation && $F$(7, 4287) \\\toprule
        Care & \includegraphics[width=.25\textwidth]{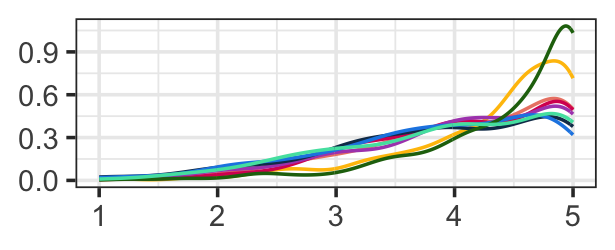} & 34.48*  \\\hline
        Equality & \includegraphics[width=.25\textwidth]{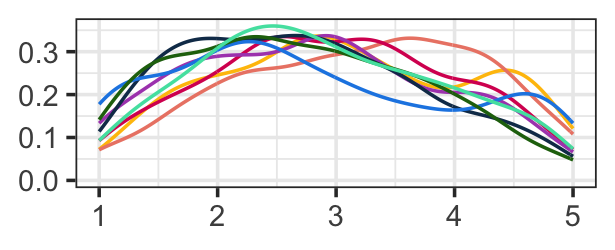} & 13.37* \\\hline
        Propor. & \includegraphics[width=.25\textwidth]{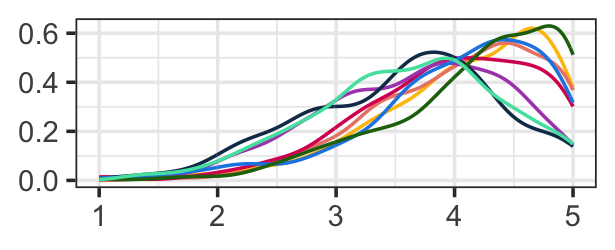} & 51.24* \\\hline
        Authority & \includegraphics[width=.25\textwidth]{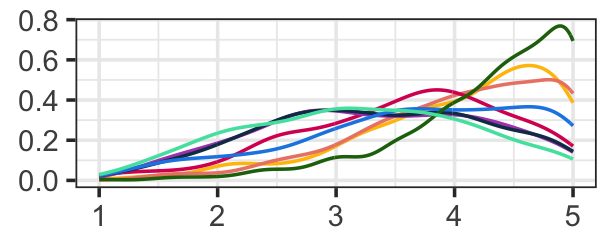} & 102.25* \\\hline
        Loyalty & \includegraphics[width=.25\textwidth]{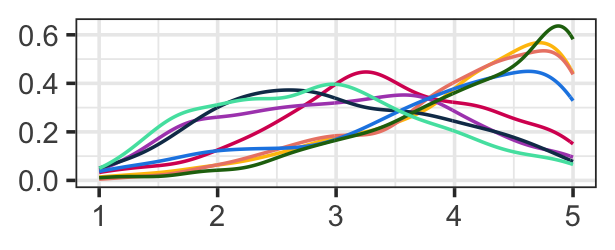} & 158.30* \\\hline
        Purity & \includegraphics[width=.25\textwidth]{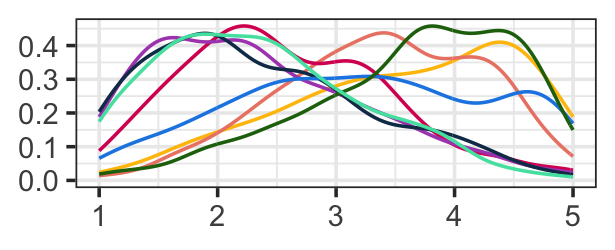} & 203.55* \\\hline
        \includegraphics[width=.45\textwidth]{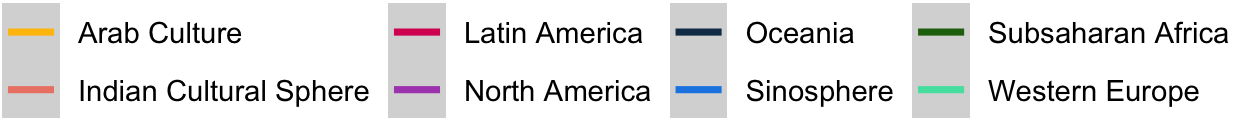}&  &  \\\hline
    \end{tabular}}
    \caption{Distribution of the moral foundations scores and the results of one-way ANOVA analysis conducted for each moral foundation across regions. *means the $p$-value of the analysis is lower than .001}
    \label{tab:moral-region}
\end{table}

\begin{figure}[t]
    \centering
    \includegraphics[width=.47\textwidth]{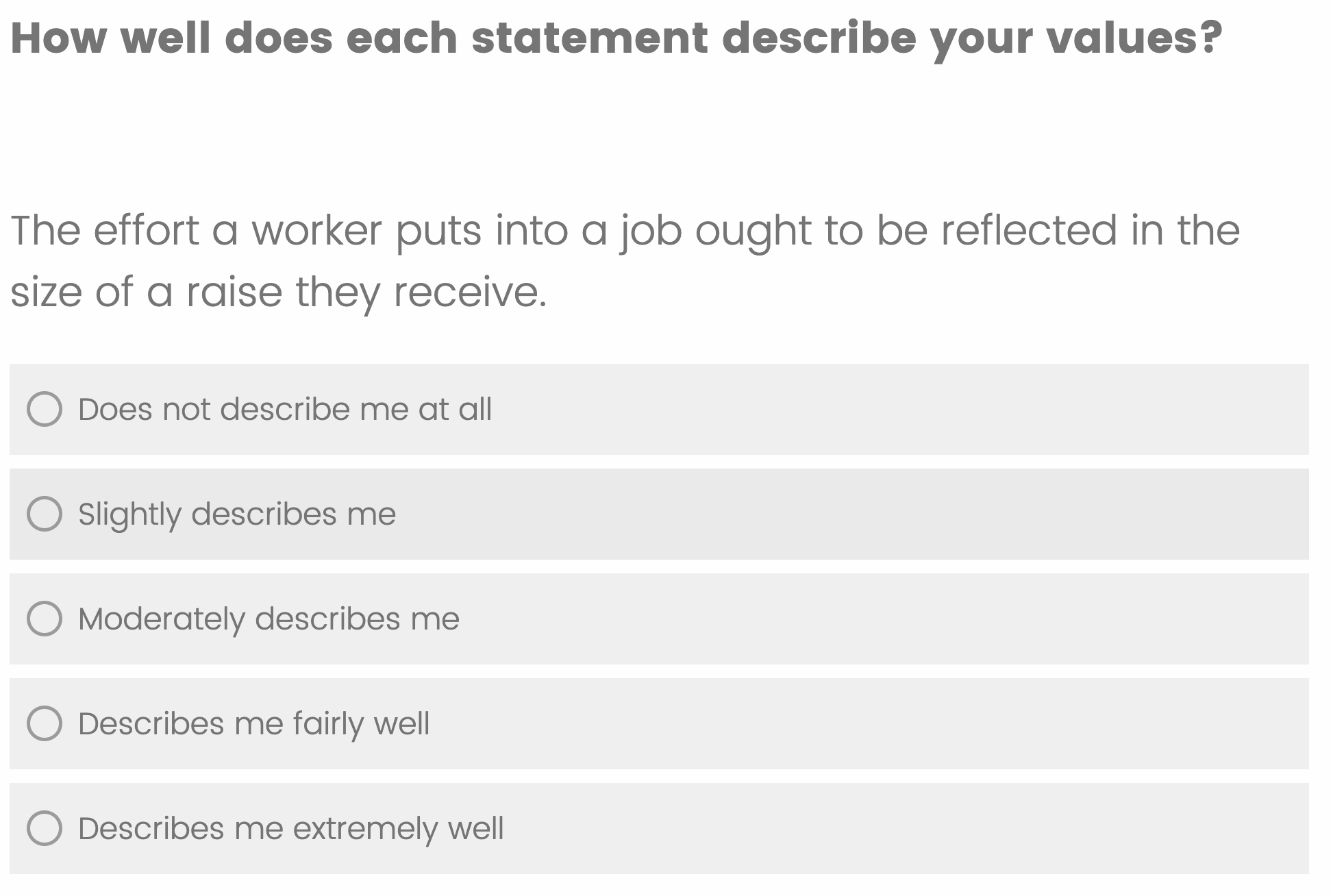}
    \caption{Sample of MFQ-2 questions in our survey}
    \label{fig:mfq}
\end{figure}

\subsection{Regions and Countries}
\label{sec:regions}
Our selected list of geo-cultural regions and countries within regions is not meant to be exhaustive, rather just to make sure that our study is done on a set of countries with diverse cultural histories. Each region listed has countries and sub-regions that have distinct cultural practices, and it is wrong to assume that the country we choose would comprehensively represent that region. Similarly, the countries listed are meant as likely places to collect data from, based on familiarity with previous data collection efforts, which potentially reflect the power structures existing within those regions. Also, each country is rarely a monolith in terms of culture (e.g., India has diverse subcultures, Australia being characterized as a ``Western'' culture erases the vibrant Australian Aboriginal culture). Data collected would also reflect the local disparities in who tends to be in the social strata that maximally overlaps with the data-collection workforce in those respective regions, and what subcultures they represent.

\begin{figure}[]
\centering
  \includegraphics[width=.47\textwidth]{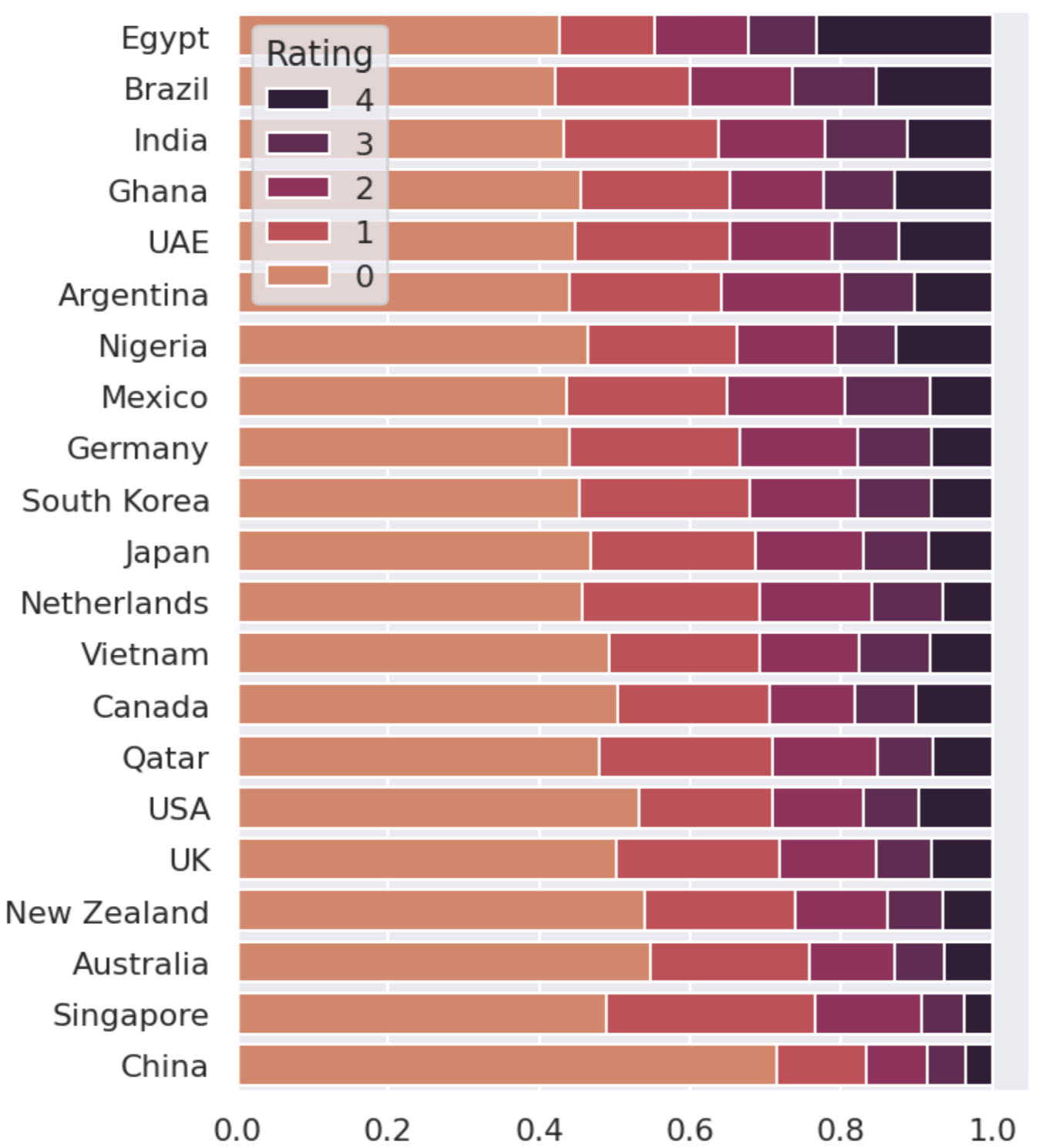}
  \caption{Distribution of the different labels provided by annotators of different countries. The y-axis is sorted based on the average offensive label captured in each country.}
    \label{fig_stats}
\end{figure}

\subsection{Test Items}
\label{sec:test}
We used participants' response to 5 test items to evaluate their understanding of offensiveness. These items (listed below) were unanimously labeled as \textit{not offensive at all} by authors and were used as an attention check to remove participants who annotated them otherwise. 

\begin{itemize}[noitemsep,nolistsep]
    \item Could you explain your point in a little more detail please.
    \item The NY Times yesterday had a rebuttal by the Iranian UN ambassador.
    \item Ditto to everything said here.
    \item Just looked at the link. Impressive and missed opportunity.
    \item Don't be so hard on yourself. Your life will go on.
\end{itemize}

\begin{table}[]
    \centering
    \scalebox{.75}{
    \begin{tabular}{ll}
    \toprule
    Region & Country \\\midrule
        Arab Culture & Egypt, Qatar, UAE \\
        Indian Cultural Sphere & India, Singapore \\
        Latin America & Argentina, Brazil, Mexico \\
        North America & Canada, USA \\
        Oceania & Australia, New Zealand \\
        Sinosphere & China, Japan, South Korea, Vietnam \\
        Sub-Saharan Africa & Ghana, Nigeria \\
        Western Europe & Germany, Netherlands, UK\\\bottomrule
    \end{tabular}}
    \caption{List of regions and countries within them in our dataset.}
    \label{tab:region-country}
\end{table}

\subsection{Data Cleaning}
We selected thresholds for the amount of time needed to finish the survey and removed annotators who performed the task either quicker or slower than the expectation. Annotators with similar answers to all items were also removed from the data.

\begin{figure}
    \centering
    \includegraphics[width=.47\textwidth]{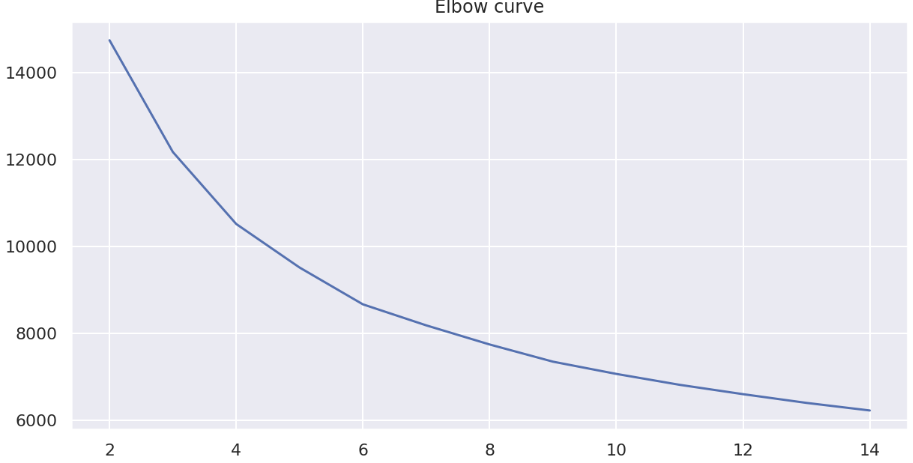}
    \caption{The distortion value captured for different options for number of moral clusters. }
    \label{fig:kmeans}
\end{figure}

\subsection{Moral clusters}
\label{sec:kmeans}

Figure \ref{fig:kmeans} shows the plot of distortions that led to us selecting 6 as the optimal number of moral clusters.

\end{document}